\documentclass[runningheads]{llncs}
\usepackage{graphicx}
\usepackage{multirow}
\usepackage{booktabs}
\usepackage{tablefootnote}
\usepackage{float}
\usepackage{tabularx}
\usepackage{soul}
\usepackage[dvipsnames]{xcolor}
\newcommand{\ds}{AuSTR}
\usepackage{blindtext}
\usepackage{soul}  
\usepackage{hyperref}
\setstcolor{red} 
\definecolor{lightblue}{rgb}{.50,.90,0.51}
\definecolor{tri}{rgb}{.25,.88,.82}
\definecolor{lilac}{rgb}{0.85,0.64,0.85}
\definecolor{atomictangerine}{rgb}{1.0, 0.6, 0.4}
\begin{document}

\title{Detecting Stance of Authorities towards Rumors in Arabic Tweets: A Preliminary Study}
\author{Fatima Haouari \and
Tamer Elsayed 
}

\authorrunning{F. Haouari et al.}

\institute{Qatar University\\
Computer Science and Engineering Department\\
\email{\{200159617,telsayed\}@qu.edu.qa}}
\maketitle             
\begin{abstract}
A myriad of studies addressed the problem of rumor verification in Twitter by either utilizing evidence from the propagation networks or external evidence from the Web. However, none of these studies exploited evidence from trusted authorities. In this paper, we define the task of detecting the stance of authorities towards rumors in tweets, i.e., whether a tweet from an authority agrees, disagrees, or is unrelated to the rumor. We believe the task is useful to augment the sources of evidence utilized by existing rumor verification systems. We construct and release the first Authority STance towards Rumors (\ds) dataset, where evidence is retrieved from authority timelines in Arabic Twitter. Due to the relatively limited size of our dataset, we study the usefulness of existing datasets for stance detection in our task. We show that existing datasets are somewhat useful for the task; however, they are clearly insufficient, which motivates the need to augment them with annotated data constituting stance of authorities from Twitter.
\keywords{Evidence \and Claims \and Social media}
\end{abstract}
\section{Introduction}

Existing studies for rumor verification in social media exploited the propagation networks as a source of evidence, where they focused on the stance of replies~\cite{wu2019different,kumar-carley-2019-tree,chen2020modeling,yu-etal-2020-coupled,bai2022multi,roy2022gdart}, structure of replies~\cite{ma2018rumor,bian2020rumor,choi2021dynamic,song2021temporally,haouari2021arcov19,bai2022rumor}, and profile features of retweeters~\cite{liu2018early}. Recently, Dougrez-Lewis et al.~\cite{dougrez2022phemeplus} proposed augmenting the propagation networks with evidence from the Web. To our knowledge, no previous research has investigated exploiting evidence for rumor verification in social media from the timelines of trusted authorities, where an authority is \emph{an entity with the real knowledge or power to verify or deny a specific rumor}~\cite{2023clef}. We believe that detecting stance of relevant authorities towards rumors can be a great asset to augment the sources of evidence utilized by existing rumor verification systems. It can also serve as a valuable tool for fact-checkers to automate their process of checking authority tweets to verify certain rumors. It is worth mentioning that stance of authorities can be just one (\emph{but} important) source of evidence that compliment other sources and by itself may not (in some cases) be fully trusted to decide the veracity of rumors.

In this paper, we conduct a preliminary study for detecting stance of authorities towards rumors spreading in Twitter in the Arab world.
Exploiting sources of evidence for Arabic rumor verification in Twitter is still under-studied; existing studies exclusively focused on the tweet text for verification~\cite{elhadad2020covid,hasanain2020overview,mahlous2021fake,al2021arabic,9620509,alqurashi2021eating}. A notable exception is the work done by Haouari et al.~\cite{haouari2021arcov19} that utilized the replies, their structure, and repliers' profile features to verify Arabic COVID-19 rumors. 
Several studies addressed Arabic stance detection in Twitter; however, the target was a specific topic not rumors~\cite{darwish2017improved,jaziriyan2021exaasc,alqurashi2022stance}. A few datasets for stance detection for Arabic claim verification were released recently, where the evidence is either news articles~\cite{baly-2018,Arastance} or manually-crafted sentences~\cite{ANS}. However, there is no dataset where the rumors are tweets and the evidence is retrieved from authority timelines, neither in Arabic nor in other languages. To fill this gap, the contribution of our work is four-fold: (1) we define the task of detecting stance of authorities towards rumors in tweets, (2) we construct and release the first Authority STance for Rumors (\ds{}) dataset,\footnote{\url{https://github.com/Fatima-Haouari/AuSTR}} (3) we present the first study on the usefulness of existing stance detection datasets for our task, and (4) we perform a failure analysis to gain insights for the future work on the task. The research question we aim to address in this work is whether the existing datasets of Arabic stance detection for claim verification are useful for detecting the stance of authorities in Arabic tweets.

The remainder of this paper is organized as follows. We outline the construction methodology of \ds{} in Section~\ref{ourData}. Our experimental setup is presented in Section~\ref{setup}. We discuss and analyze our results in Section~\ref{discuss}. Finally, we conclude and suggest some future directions in Section~\ref{conclusion}.

% Arabic stance studies where the target is a topic: 
% 1) towards distance education in the region of Saudi Arabia \cite{alqurashi2022stance} (did not release data). 
% 2) \cite{darwish2017improved} studied the stance of tweets towards a single topic "the transfer of ownership of the islands of Tiran and Sanafir
% from Egypt to Saudi Arabia" or "ISIS attacks on Paris"3) \cite{mubarak2022arcovidvac} release an Arabic Twitter datasets with stance towards COVID19 vaccination.Arabic target-based stance dataset \cite{jaziriyan2021exaasc}, they annotated replies towards a target in the source tweet. (released data)cross-domain~\cite{hardalov-etal-2021-cross}
% cross-lingual stance detection~\cite{hardalov2022few}
% In our work we assume that the system have a rumor expressed in a tweet, and a set of authorities Twitter accounts, and the task is detect the stance of the authorities tweets toward the given rumor.

% \begin{enumerate}
%     \item The first study addressing evidence detection from authorities Twitter accounts for rumor verification.
%     \item We study the usefulness of existing stance for claim verification datasets for evidence detection from authorities.
%     \item We manually construct and release a test set of rumor-evidence pairs from authorities in Twitter.  
%     \item We perform an extensive error analysis to study to gain insights for the future improvements of models and datasets for evidence retrieval from authorities.
% \end{enumerate}
\section{Constructing \ds{} Dataset
%: A New Test Dataset for Stance of Authorities
}\label{ourData}

% Based on the literature review, we found that there is no study addressing detecting the stance of authorities towards rumors in tweets in general and in Arabic specifically. Thus, 
% \hl{why we need such dataset?(for test) objectives/criteria?}
% To construct our new Authority STance for Rumors dataset (\ds),
To construct \ds{} 
where both the rumor and evidence are tweets, we exploit both fact-checking articles and variant authority Twitter accounts.% covering multiple domains and countries. 

\paragraph{\textbf{Exploiting Fact-checking Articles.}} 
Fact-checkers who attempt to verify rumors usually provide in their fact-checking articles some examples of social media posts (e.g., tweets) propagating the specific rumors, and other posts from trusted authorities that constitute evidence to support their verification decisions. To construct \ds, we exploit both examples of tweets: stating rumors and showing evidence from authorities as provided by those fact-checkers. Specifically, we used AraFacts~\cite{ali2021arafacts}, a large dataset of Arabic rumors collected from 5 fact-checking websites. From those rumors, we selected only the ones that are expressed in tweets and have evidence in tweets as well.\footnote{We contacted the authors of AraFacts to get this information as it was not released.} We then extracted the rumor-evidence pairs as follows. For \textit{true} and \textit{false} rumors, we selected a single tweet example and all provided evidence tweets, which are then labeled as having \textit{agree} and \textit{disagree} stances respectively.\footnote{We only kept evidence expressed in \emph{text} rather than in image or video.} 
%We then manually checked each pair to make sure that both the rumor and the evidence are in Arabic,\footnote{Non-Arabic rumor and evidence are provided in some cases}
If the fact-checkers provided the authority account but stated no evidence was found to support or deny the rumor, we selected one or two tweets from the authority timeline posted soon before the rumor time, and assigned the \textit{unrelated} label to the pairs.

\paragraph{\textbf{Exploiting Authority Accounts}.} Given that fact-checkers focus more on \textit{false} rumors than \textit{true} ones, we ended up with only 4 \textit{agree} pairs as opposed to 118 \textit{disagree} pairs following the above step. To further expand our \textit{agree} pairs, we did the reverse of the previous approach, where we collected the evidence first. Specifically, we started from a set of Twitter accounts of authorities (e.g., ministers, presidents, embassies, organization accounts, etc.) covering most of the Arab countries and multiple domains (e.g., politics, health, and sports), and selected recent tweets stating claims from their timelines. For each claim, we used Twitter search interface to look for tweets from regular users expressing it, but tried to avoid exact duplicates.
Finally, to get closer to the real scenario, where percentage of \textit{unrelated} tweets is usually higher than percentages of \textit{agree} and \textit{disagree} tweets in the authority timelines, we further expanded the \textit{unrelated} pairs by selecting one or two \textit{unrelated} recent tweets from the authority timeline posted before the rumor time for each \textit{agree} and \textit{disagree} pairs. 

% \begin{enumerate}
%     \item We selected a set of authorities Twitter accounts such as ministers, presidents, embassies, organizations, etc. Covering most of the Arabic countries, and multiple domains including politics, health, sports to name a few. We then selected recent tweets stating claims from their timeline.
%   \item We used the Twitter search interface to look for tweets from regular users expressing the claim but trying to avoid exact duplicates.
   
% \end{enumerate}

Overall, we end up with 409 pairs covering 171 unique claims, where 41 are \textit{true} and 130 are \textit{false}. Among those pairs, 118 are \emph{disagree} (29\%), 62 are \emph{agree} (15\%), and 229 are \emph{unrelated} (56\%).
\section{Experimental Setup}\label{setup}
%In this section we present our experimental setting in terms of training data and stance models used in our experiments.
%We describe the training data used in our experiments in Section~\ref{trainingData}, and present our stance models and their setup in Section~\ref{stanceModels}. 

%\subsection{Stance training data}\label{trainingData}
\paragraph{\textbf{Datasets}.} 
%Due to the limited size of our data, and 
To study the usefulness of existing Arabic datasets that target stance for claim verification, we adopted the following ones for training:
\begin{enumerate}
    \item \textbf{ANS~\cite{ANS}} of 3,786 \textbf{(claim, sentence)} pairs, where claims were extracted from news article titles from trusted sources, then annotators were asked to generate \textit{true} and \textit{false} sentences towards them by adopting paraphrasing and contradiction respectively. The sentences are annotated as either \textit{agree}, \textit{disagree}, or \textit{other} towards the claims.
    \item \textbf{ArabicFC~\cite{baly-2018}} of 3,042 \textbf{(claim, article)} pairs, where claims are extracted from a single fact-checking website verifying political claims about war in Syria, and articles collected by searching Google using the claim. The articles are annotated as either \textit{agree}, \textit{disagree}, \textit{discuss}, or \textit{unrelated} to the claim. 
    \item \textbf{AraStance~\cite{Arastance}}: 3,676 \textbf{(claim, article)} pairs, where claims are extracted from 3 Arabic fact-checking websites covering multiple domains and Arab countries. The articles were collected and annotated similar to ArabicFC.
\end{enumerate}

To train our models, we considered only three labels, namely, \textit{agree}, \textit{disagree}, or \textit{unrelated}. For ANS and AraStance, we used the same data splits provided by the authors; however, we split the ArabicFC into 70\%, 10\%, and 20\% of the claims for training, development, and testing respectively\footnote{We release ArabicFC splits for reproducibility.}. When splitting data, we assigned all pairs having the same claim to the same split. Table~\ref{tab:train-data} shows the size of different data splits of the three datasets. Due to the limited size of \ds, in this work, we opt to utilize it only as a \emph{test set} while using the above datasets for training to show their usefulness in our task. 

\begin{table}[h]
\caption{Data splits of the Arabic stance datasets used for training.}
\label{tab:train-data}
\centering
\begin{tabular}{c|ccc|ccc|ccc}
\multirow{2}{*}{\textbf{Label}} & \multicolumn{3}{c|}{\textbf{ANS}} &\multicolumn{3}{c|}{\textbf{ArabicFC}} & \multicolumn{3}{c}{\textbf{AraStance}}                                    \\  & \textbf{Train}  & \textbf{Dev}  & \textbf{Test} & \textbf{Train} & \textbf{Dev} & \textbf{Test}  & \textbf{Train} & \textbf{Dev} & \textbf{Test}    
\\
\hline
\textbf{Agree} & 903 & 268 & 130 &323 & 32 & 119 & 739 & 129 & 154 \\
\textbf{Disagree}& 1686& 471& 242 & 66 & 8& 13 & 309 & 76 & 64 \\
\textbf{Unrelated}  & 63 & 16 & 7  &1464&198&410 & 1553 & 294 & 358 \\
\hline 
Total & 2652& 755 & 379 & 1853& 238&542& 2601 &499 & 576 \\
\bottomrule
\end{tabular}
\end{table}

\paragraph{\textbf{Stance Models}.}\label{stanceModels}
To train our stance models, we fine-tuned BERT~\cite{devlin2018bert} to classify whether the evidence sentence/article \textit{agrees} with, \textit{disagrees} with, or is \textit{unrelated} to the claim. We feed BERT the claim text as sentence \emph{A}, the evidence as sentence \emph{B} (truncated if needed) separated by the [SEP] token. Finally, we use the contextual representation of the [CLS] token as input to a single classification layer with three output nodes, added on top of the BERT architecture to compute the probability for each class of stance.

Various Arabic BERT-based models were released recently~\cite{antoun2020arabert,safaya-etal-2020-kuisail,lan-etal-2020-empirical,inoue2021interplay,abdul2021arbert}; we opted to choose ARBERT~\cite{abdul2021arbert} as it was shown to achieve better performance on the stance datasets adopted in our work~\cite{Arastance}. We adopted the authors' setup~\cite{Arastance} by training the models for a maximum of 25 epochs, where early stopping was set to 5 and sequence length to 512. We trained 7 different models in an ablation study using different combinations of the stance datasets mentioned earlier.

\section{Results and Discussion}\label{discuss}
%\subsection{Evaluation results}
The research question we address in this preliminary study is whether the existing stance detection datasets are useful or not in our task. To answer it, we use combinations of the existing datasets for training and \ds{} for testing. We also show how models trained on those combinations perform on their own corresponding in-domain test sets. While the results on the in-domain test sets are not comparable, since those test sets are different, they constitute an estimated upper bound performance. To evaluate the models, we report per-class $F_1$ and macro-$F_1$ scores. % (to overcome the imbalance nature of all the test sets)
Table~\ref{tab:results} presents the performance results of all experiments, which demonstrate several interesting observations. 

%AuSTR is more challenging: performance is much lower than in-domain
%Disagree is the most challenging -- best performance is in sixties of F1

First, we notice that almost all models (except a few) were able to achieve higher performance on their own in-domain test sets compared to \ds. This shows that domain adaptation was not very effective (thus in-domain data for our task is required for training the models).

Second, when using individual stance datasets for training, the model trained on AraStance clearly outperformed the others in all measures when tested on \ds. We note that ArabicFC is severely imbalanced, where the \textit{disagree} class represents only 3.3\% of the data, yielding a very poor performance on that class even when tested on its own in-domain test set. A similar conclusion was found by previous studies~\cite{baly-2018,Arastance}. 
%Moreover, it covers only political claims focusing on Syrian war. 
As for ANS, evidence is manually crafted, which is not as realistic as tweets from authorities. Alternatively, AraStance claims are extracted from three fact-checking websites,\footnote{Claims are collected from sources other than the ones we used to construct \ds.} covering multiple domains and Arab countries, similar to \ds, and the evidence is represented in articles written by journalists, not manually crafted. 

Third, when tested on \ds, the model trained on all datasets combined exhibits the best performance on the \emph{disagree} class; however its performance was severely degraded compared to the AraStance model on the \emph{agree} class.
%, with a very similar performance on the \emph{unrelated} class. 
This indeed needs further investigation.

%Second, when combining two datasets for training, combining AraStance and ArabicFC yielded better performance than the AraStance model only for the \emph{disagree} class when tested on \ds. We note that both share the same domain of fact-checking articles. In fact, performance was degraded severely for the \emph{agree} class. However, the model trained on all datasets combined exhibits the best overall performance in all per-class measures except on the \emph{agree} class when tested on \ds. This indeed needs further investigation.

%where the context are articles and target are claims from fact-checking websites. Another observation is that
%although ArabicFC and ANS achieved the lowest performance in terms of average-F1 when used separately for training, when combined the performance improved significantly. This can be attributed to the fact that ANS and ArabicFC are severely imbalanced in terms of \textit{unrelated} and \textit{disagree} classes respectively, so combining them yielded to a better balanced data. This is clear also on the model performance on the data own test set. 

%for the models trained on ArabicFC and AraStance+ArabicFC where they achieved a higher $F_1$-score on the \textit{disagree} class on \ds{}.

Furthermore, we observe that AraStance achieved the highest $F_1$(D) when used solely for training, and whenever combined with the other datasets. To investigate this, we manually examined a 10\% random sample of \textit{disagreeing} training articles. We found they have common words such as \textit{rumors}, \textit{not true}, \textit{denied}, and \textit{fake}; similar keywords appear in some \textit{disagreeing} tweets of \ds. %This is in contrast to ANS, where \textit{disagreeing} sentences are constructed by modifying claim terms and avoiding explicit negation. 
%The results also show that the models trained with AraStance solely achieved the highest $F_1$(A) compared to all other models. We believe this can be attributed to the fact that most of AraStance True claims where collected from Reuters which are basically news about countries and authorities, similarly \ds{} True claims were collected from authorities timeline where we may have an overlap in the writing style,\footnote{Based on exploring a sample of the agreeing articles in the train set} e.g., \textit{The ministry announced/confirmed, The army reported, The Secretary General said}, etc. 

% Furthermore, it remains unclear why AraStance achieved the highest $F_1$-score on the \textit{agree} class, while when combined with other datasets, the performance on that class degraded significantly. We believe this needs further investigations.

Finally, we observe that there is a clear discrepancy in the performance across different classes. Considering the model trained on all datasets for example, $F_1$(A) is 0.74 while $F_1$(D) is 0.65. Moreover, it is clear that detecting the \textit{disagree} stance is the most challenging subtask, which we expect to benefit from in-domain training. Overall, we believe training and testing on tweets is very different, as they are very short and informal, which needs special pre-processing.

% To further get insights for possible future improvements to datasets for detecting stance of authorities, we present our failure analysis in the next section. 
% Finally, it is clear that detecting the \textit{disagree} stance is the most challenging task. 

% To further get insights for possible future improvements to datasets for detecting stance of authorities, we present our failure analysis in the next section. 

\begin{table*}
\caption{Performance on both the in-domain test sets and \ds{}, measured in per-class $F_1$ (A: Agree, D: Disagree, U: Unrelated) and macro-$F_1$. On \ds, bold and underlined values indicate best and second-best performance respectively.}
\label{tab:results}
\centering
\begin{tabular}{l||ccc|c||ccc|c}
\toprule
 & \multicolumn{4}{c||}{\textbf{Test on in-Domain Set}} 
& \multicolumn{4}{c}{\textbf{Test on \ds}} 
\\
\textbf{Training Set} & $F_1$(A) & $F_1$(D) & $F_1$(U) 
% & \textbf{Acc}
& m-$F_1$ & $F_1$(A) & $F_1$(D) & $F_1$(U) &
% \textbf{Acc} &
m-$F_1$ \\
\hline
ANS & \,0.824\,     & \,0.901\,      & \,0.923\,      
% & 0.873 
& \,0.882\,      & \,0.653\,      & \,0.578\,      & \,0.709\,
% & 0.646
& \,0.647\,       \\
ArabicFC&     \,0.770\,  &   \,0.090\,     &\,0.915\,      
% &        0.862 
&  \,0.591\,       &  \,0.641 \,   &\,0.434\,      &\,0.799\,  
% &0.671
&\,0.625\,       \\
AraStance& \,0.898\,      & \,0.833\,     & \,0.95\,       
% & 0.924  
& \,0.894\,       & \,\textbf{0.837}\,      & \,0.613\,      & \,\underline{0.865}\,      
% & \textbf{0.795}    
& \,\textbf{0.772}\,
\\
\hline
ANS+ArabicFC             &\,0.807\,   &    \,0.866\,   & \,0.899\,     
% &    0.863
&   \,0.857\,   &  \,0.678\, &  \,0.587\,    &\,0.862\,   
% &0.756  
&\,0.709\,
\\
ANS+AraStance& \,0.893\,  &  \,0.909\,     &    \,0.955\,  
% & 0.921  
& \,0.919\,     & \,0.743\,     & \,0.629\,  &   \,0.847\,   
% &    0.768  
& \,0.740\,
\\
ArabicFC+AraStance & \,0.765\, &    \,0.555\,    &   \,0.897\,   
% &      0.83  
&  \,0.739\,   &      \,\underline{0.754}\, & \,\underline{0.635}\,  &  \,0.862\,    
% & 0.783    
& \,0.750\,
\\
\hline
All Three Datasets\,              &   \,0.778\, &   \,0.742\,    &\,0.889\,      
% &      0.824  
&  \,0.803\,   &\,0.741\,      &  \,\textbf{0.646}\,  &   \,\textbf{0.866}\,      % &\underline{0.793}  
& \,\,\underline{0.751}\,\, \\
\bottomrule
\end{tabular}
\end{table*}

%\subsection{Failure analysis}
\paragraph{\textbf{Failure Analysis}.}
We conducted a failure analysis on 17 examples from \ds{} that failed to be predicted correctly by \textit{all} of our 7 trained models. We found that we can attribute the failures to two main reasons: (1) \textit{Writing Style}, where the authority is denying a rumor about herself speaking in the first person. This constitutes 64.7\% of the examined failures. We believe this is due to the fact that none of the stance datasets we used for training have evidence written by authorities themselves, as the source was either news articles written by journalists, or paraphrased or contradicted news headlines manually crafted by annotators. (2) \textit{Indirect Disagreement/Agreement}, where the authority is indirectly  denying/supporting the rumor. Examples of both types of failures are presented in Table~\ref{tab:failed_examples}. These findings motivate the need to augmenting existing stance datasets with rumor-evidence pairs from Twitter to further improve the performance of detecting the stance of authorities towards rumors from their tweets.

% We found that generally when the authority himself tweeted and is denying a false rumor about himself, the models failed to detect the stance. We believe this is attributed to the fact that none of the stance datasets we used for training have evidence written by authorities themselves as the evidence are either news articles written by journalists, or paraphrased or contradicted news headlines manually crafted by annotators. The writing style is different. Moreover, none of them is a Twitter dataset. Moroever, as we notice in the evidence tweets in the first 2 rows of Table~\ref{tab:failed_examples}, although the evidence tweet is posted by authorities the rumor is about,there is nothing in the text that shows this is written by them which justifies why a model relying on text is detecting that the evidence is unrelated. 

% These finding motivate the need to first, exploring expanding the evidence with extra context such the authority name and profile description, and experiment with other models not depending on text solely. Second, augment existing stance datasets with rumor-evidence pairs from Twitter to further improve the performance of stance of authorities.  

\begin{table}[h]
\caption{Sample examples failed to be predicted correctly by \underline{\textbf{all}} models. The golden label for the examples is either \color{ForestGreen} Agree \color{black} or \color{red}Disagree\color{black}. Failure types are writing style, indirect disagreement, and indirect agreement for the examples in order.}\label{tab:failed_examples}
\centering
\begin{tabular}{p{5.cm}@{\hskip 0.1in}p{7.cm}@{\hskip 0.1in}}
\hline
\textbf{Rumor tweet [posting date]} &  \textbf{Evidence tweet [posting date]}\\
\hline
\texttt{Mortada Mansour passed away recently of a heart attack.[29-10-2021]}& \textbf{@Mortada5Mansour}:\color{red}\texttt{ I am having my dinner now, and after a few minutes I will share a voice and video to reassure you, and I will reply to those who disturbed my family members in my village and caused the anxiety to all my fans.[29-10-2021]}\\

\hline
\texttt{Egypt does not give a vaccine to its citizens, the Gulf countries sponsor them: Saudi Arabia / Sultanate of Oman / Qatar refuses their intervention, so there is no other than Kuwait, the country of humanity that receives them and feeds them. What is the mysterious secret? Kuwait treats Egypt with special treatment.[07-05-2021]} & \textbf{@mohpegypt}:\color{red}\texttt{ Information about the \#coronavirus vaccine. To book a vaccine, please visit the website http://egcovac.mohp.gov.eg
or go to the nearest health unit (for citizens who have difficulty registering online). For more information, please call the hotline: 15335 \#together\_rest\_assured.[10-05-2021] }
\\
\hline

\texttt{Urgent
The headquarters of the fourth channel was stormed by the militias of the Sadrist movement in the capital, Baghdad.[04-11-2022]} & \textbf{@MAKadhimi}:\color{ForestGreen}\texttt{ The attack on one of the Iraqi media outlets, and the threat to the lives of its employees, is a reprehensible act and represents the highest level of transgression against the law and freedom of the press and does not fall within the peaceful and legal practices and protests.
We directed that the perpetrators be held accountable, and that protection be tightened on press institutions.[04-11-2022] }
\\
 
\bottomrule
\end{tabular}
\end{table}
\section{Conclusion and Future Work}\label{conclusion}
In this paper, we defined the task of detecting stance of authorities towards rumors in tweets, and released the first dataset for the task targeting Arabic rumors. We studied the usefulness of existing Arabic datasets for stance detection for claim verification in our task. Based on our experiments and failure analysis, we found that although existing stance datasets showed to be somewhat useful for the task, they are obviously insufficient and there is a need to augment them with stance of authorities from Twitter data. In addition to expanding \ds{} to have sufficient training data for the task that can be used solely or to augment existing stance datasets, we plan to explore and contribute with stance models specific to the task.
\section*{Acknowledgments} The work of Fatima Haouari was supported by GSRA grant\# GSRA6-1-0611-19074 from the Qatar National Research Fund. The work of Tamer Elsayed was made possible by NPRP grant\# NPRP11S-1204-170060 from the Qatar National Research Fund (a member of Qatar Foundation). The statements made herein are solely the responsibility of the authors.
\bibliographystyle{splncs04}
\bibliography{references}
\end{document}